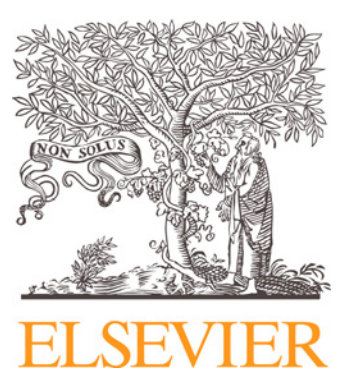

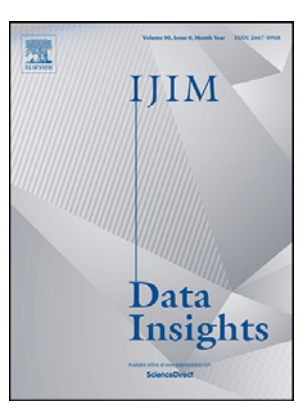

# Deep learning in business analytics: A clash of expectations and reality

Marc Schmitt[a,b]

[a] *Department of Computer Science, University of Oxford, UK*
[b] *Department of Computer and Information Sciences, University of Strathclyde, UK*

ARTICLE INFO

ABSTRACT

Our fast-paced digital economy shaped by global competition requires increased data-driven decision-making based on artificial intelligence (AI) and machine learning (ML). The benefits of deep learning (DL) are manifold, but it comes with limitations that have – so far – interfered with widespread industry adoption. This paper explains why DL – despite its popularity – has difficulties speeding up its adoption within business analytics. It is shown that the adoption of deep learning is not only affected by computational complexity, lacking big data architecture, lack of transparency (black-box), skill shortage, and leadership commitment, but also by the fact that DL does not outperform traditional ML models in the case of structured datasets with fixed-length feature vectors. Deep learning should be regarded as a powerful addition to the existing body of ML models instead of a "one size fits all" solution. The results strongly suggest that gradient boosting can be seen as the go-to model for predictions on structured datasets within business analytics. In addition to the empirical study based on three industry use cases, the paper offers a comprehensive discussion of those results, practical implications, and a roadmap for future research.

## 1. Introduction

The last decade was shaped by huge improvements in data storage and analytics capabilities (Baesens, Bapna, Marsden, Vanthienen, & Zhao, 2016; Henke et al., 2016). What started as the big-data (Kushwaha, Kar, & Dwivedi, 2021) revolution brought us the age of constant digital change, accelerating globalization, and the continuous move toward a digital world economy (Davenport, 2018; Warner & Wäger, 2019). Companies operating in today's world have to deal with global competition in an ultra-fast marketplace (Davenport, 2018), and AI-enabled information management (Borges, Laurindo, Spínola, Gonçalves, & Mattos, 2021; Collins, Dennehy, Conboy, & Mikalef, 2021; Duan, Edwards, & Dwivedi, 2019; Verma, Sharma, Deb, & Maitra, 2021) is the key to navigating the digital storm of the 21st century.

Artificial intelligence (AI) and machine learning (ML) have been widely accepted as general-purpose technology for decision-making (Agrawal, Gans, & Goldfarb, 2019) across a variety of domains, industries, and functions including biotech, healthcare (Sounderajah et al., 2022; Young & Steele, 2022), marketing (Verma et al., 2021), human resource management (Votto, Valecha, Najafirad, & Rao, 2021), financial services (Schmitt, 2020; Singh, Chen, Singhania, Nanavati, & Gupta, 2022), insurance (Rawat, Rawat, Kumar, & Sabitha, 2021), risk management (Fujii, Sakaji, Masuyama, & Sasaki, 2022; Schmitt, 2022b), cybersecurity (Taddeo, McCutcheon, & Floridi, 2019; Thorat, Parekh, & Mangrulkar, 2021), and many others (Kumar, Kar, & Ilavarasan, 2021). Data analytics and information systems (Kushwaha et al., 2021) build the foundation for actionable insights and are the primary enablers for AI-based decision-making across all domains.

The function responsible for converting raw data into valuable business insights is called business analytics. It is an interdisciplinary field drawing and combining expertise from machine learning, statistics, information systems, operations research, and management science (Sharda, Delen, & Turban, 2017). Business analytics constitutes a quite long chain of different analytics, which includes descriptive, predictive, and prescriptive analytics (Delen & Ram, 2018). ML operates mainly in the predictive sphere of business intelligence but has started to incorporate prescriptive analytics as well (Bertsimas & Kallus, 2019).

One of the major technologies responsible for driving the current digital revolution (Agrawal et al., 2019; Bughin et al., 2017) is deep learning (LeCun, Bengio, & Hinton, 2015). It is a subset of ML and emerged out of earlier research on brain-inspired neural networks. DL is capable of learning complex hierarchical representations of data. It was able to outperform traditional methods and has predictive capabilities that come close to or surpass human-level intelligence in different areas. The main reasons for the breakthrough of DL stem from developments in three different areas (Goodfellow, Bengio, & Courville, 2016): (1) Optimization algorithms allow the training of deep neural networks (Hinton, Osindero, & Teh, 2006); (2) The era of "big data" increased the amount of large structured, as well as unstructured data-sets, which are now ripe for harvesting (Chen, Chiang, & Storey, 2012; Kushwaha et al.,

*E-mail address:* marcschmitt@hotmail.de

2021); and (3) hardware improvements, especially GPU's made it possible to train those highly power-hungry models with those huge datasets. Accurate performance for unstructured high-dimensional data sets became only possible due to the advances of DL, which significantly enhances the field of machine learning (Jordan & Mitchell, 2015) to tackle further use cases and take over tasks that were initially only reserved for humans (Agrawal et al., 2019).

However, there seems to be a certain confusion when it comes to the adoption of deep learning in business analytics and information management. Hence this paper is an attempt to bring clarity towards why DL might be used or not used for certain business uses cases, and what the reasons are, and also gives recommendations on where to apply DL in practice.

Most analytics departments across the corporate value chain have traditionally been using predictive statistics and machine learning models such as GLMs, CART, and ensemble learning. Those models are vital tools to help with several analytics tasks that directly impact the bottom line of firms and organizations (Siebel, 2019). Also, we have moved from fundamental progress in AI to the application of deep learning in various sciences, businesses, and governments (Lee, 2018; Stadelmann et al., 2018). Despite the huge success of DL, a closer investigation of the current literature reveals that the adoption rate for DL in business functions for analytic purposes is quite low.

Chui et al. (2018) analyzed 100 use cases to demonstrate the current deployment of AI/DL-related models across industries and business functions compared to other models referred to as traditional analytics. The result is that while the adoption of DL starts to increase, it seems most units remain working with the older more established analytical models that have been successful already years ago. McKinsey (Chui et al., 2018) also distinguishes departments that have traditionally been using analytics as compared to departments that are foreign to quantitative decision enablers. McKinsey draws a clear picture that shows that the only areas where DL has been utilized so far are traditional analytics arms that have the natural capabilities and skillsets in place to work with modern AI, while technology foreign departments are reluctant to adopt DL models. But even in business units with traditionally strong links to analytics – like risk management and insurance – the utilization of DL remains relatively low and traditional models are still the go-to solution.

Deep learning is on the way to becoming the industry standard and is broadly perceived as general-purpose technology for decision-making, however, business analytics is still in its infancy when it comes to adopting this technology. DL does not prevail within business analytics functions as perceived due to the current hype and job descriptions (Kraus, Feuerriegel, & Oztekin, 2019).

The main issues why it is not easy to develop and deploy DL – especially for small to medium-sized corporations – can be partially mapped to the three reasons why DL found its breakthrough in recent years. The following bottlenecks could be identified when it comes to the adoption of DL in business analytics functions:

(1) **Computational Complexity:** The hardware necessary to train and validate DL models on large datasets is tremendous, which makes infrastructure investments quite expensive. This stands in large contrast to the question of whether the development and implementation of those models will materialize and be reflected in a future value increase (Bughin et al., 2017).
(2) **Infrastructure:** Companies need to be able to harvest a continuous flow of unstructured data to capture the value from DL, which is difficult if the necessary "big data" infrastructure is not in place (Bughin et al., 2017).
(3) **Transparency:** Another reason is the nature of DL itself. DL is mainly a black box, which means it can predict correctly, but we lack a causal explanation of why it arrives at a certain decision (Samek & Müller, 2019). This makes it problematic for industries, which are subject to regulatory supervision.
(4) **Skill Shortage:** Talent (Kar, Kar, & Gupta, 2021) is required to implement those models as well as subject matter expertise to define use cases (Henke et al., 2016). The current supply and demand gap for ML experts makes it difficult for small- and medium-sized corporations to utilize advanced AI.
(5) **Leadership Commitment:** Full management support to establish and drive a company-wide AI strategy is also a vital prerequisite for increased adoption speed (Kar et al., 2021).

Many studies about the adoption of DL in business analytics seem to ignore its general value contribution, which should come in the form of improved prediction accuracy. DL must make a business case for itself to justify its adoption, but this is not always given. Also, complexity and infrastructure justifications cannot be the complete picture as resources are increasing consistently, powerful processors and databases do exist, and once a model is trained, the resource requirements are not that significant anymore. Another reason why DL might be lacking in certain areas could be its ability to outperform existing AI/ML models.

Several standalone studies comparing the predictive ability of deep learning against traditional machine learning methods on structured data sets have concluded that DL does not outperform tree-based ensembles (Addo, Guegan, & Hassani, 2018; Hamori, Kawai, Kume, Murakami, & Watanabe, 2018). This stands in contrast to the claim that DL offers performance improvements across the board as indicated by Kraus et al. (2019) and also to the general assumption that DL needs to be adopted in every business function (Chui et al., 2018). While the success of DL for unstructured data problems such as image recognition and NLP is beyond doubt, the reality of DL for structured data within companies' business analytics functions is less clear and is the focus of this article. Structured data with fixed-length feature vectors are vastly present in relational databases and standard business uses cases.

This paper investigates the following two research questions:

- RQ1: Does DL outperform traditional ML models for supervised learning problems in the case of structured data with fixed-length feature vectors?
- RQ2: Is Deep Learning – despite its popularity – always the right AI/ML model within business analytics and information management?

The core contribution of the paper is to paint a clear picture of deep learning in business analytics and information management in terms of its performance on structured datasets. Comments such as "DL can be a simple replacement of traditional models" are too general and not always true. For structured data, tree-based ensembles as gradient boosting seem to be at least on par with DL across different domains. In support of this claim, an empirical test using three case studies based on real-world data is presented.

Concrete, this paper will contribute to the current body of literature in the following ways:

(1) DL is compared to traditional machine learning models such as GLMs, random forest, and gradient boosting based on three real-world use cases within the context of business analytics to verify the assumption that DL does not outperform traditional methods on structured datasets.
(2) Comprehensive discussion based on the results of the empirical study including practical implications for researchers and professionals.
(3) In the end, a roadmap for future research directions to further integrate AI/ML with business analytics and information management is presented.

This article is structured as follows: Section 2 introduces the machine learning models used in this study - logistic regression, random forest, gradient boosting, and deep learning. Second, the experimental design is presented, which includes an explanation of the dataset, preprocessing steps, and the software setup. In Section 3, the numerical results from the three case studies based on real-world data/business problems are presented. All three case studies show that in the case of structured data (tabular data) DL does not have a performance advantage over the tree-based ensembles random forest and gradient boosting machine. Section 4 discusses the technical implications of these results, implications for practice, and future research directions, while Section 5 concludes with a summary.

## 2. Methods and materials

### 2.1. Machine learning

This part gives an overview of predictive analytics and the ML models used in the experiment. The ML models used and compared in this experiment are Logistic Regression (LR), Random Forest (RF), Gradient Boosting Machine (GBM), and Deep Learning (DL). For a comprehensive treatment of the underlying theory, it is referred to (Hastie, Tibshirani, & Friedman, 2017) and (Murphy, 2012) for ML and (Goodfellow et al., 2016) for DL.

#### 2.1.1. Logistic regression

The **Logistic Regression (LR)** belongs to the big family of generalized linear models (GLMs). GLMs are characterized by taking as input a linear combination of features and linking them to the output with the help of a function where the output has an underlying exponential probability distribution like the normal distribution or the binomial distribution (Murphy, 2012). The LR is the standard method for binary classification and is widely used in academia and industry. A linear combination of inputs and weights is calculated and applied by feeding $w^T x$ into the logic or sigmoid function represented by

$$sigm\left(w^T x\right) = \frac{1}{1+ e^{-w^T x}} = \frac{e^{w^T x}}{e^{w^T x}+1}. \quad (1)$$

The sigmoid function restricts the range of the output to be in the interval [0, 1].

#### 2.1.2. Random forest

The recursive partitioning algorithms **Random Forest (RF)** is part of the family of ensemble methods and operates very similar to decision trees with bagging. Bagging (Breiman, 1996) chooses randomly different M subsets from the training data with replacement and averages these estimates. The random forest creates different decision trees and averages the results in the end to reduce the variance of the prediction model (Murphy, 2012). It is one of the most potent ML algorithms for classification and regression tasks out there.

#### 2.1.3. Gradient boosting

Boosting is like bagging but builds models in a sequential order instead of averaging different results. The idea of boosting is to start with a weak learner that gradually improves by correcting the error of the previous model at each step. This process improves the performance of the weak learner and moves gradually towards higher accuracy. The most common model used for boosting is a decision tree. There are several different **Gradient Boosting (GM)** implementations out there. This paper uses the gradient boosting version implemented by (Malohlava & Candel, 2019) which is based on (Hastie et al., 2017). Gradient boosting is one of the strongest prediction models for structured data currently available.

#### 2.1.4. Deep learning

Deep Learning comes with many architectures such as feed-forward artificial neural networks (ANN), Convolutional neural networks (CNNs), as well as Recurrent Neural Networks (RNNs). The best architecture for transactional (tabular) data, which are not sequential – as in this study – is a multi-layer feedforward artificial neural network. Other, more complex architectures such as RNNs do not possess any advantage in those cases (Candel & LeDell, 2019). The architectural graph of a feed-forward neural network can be seen in Fig. 1. The first column represents the input features and is called the input layer. The last single neuron represents the output to where the final activation function is applied to. The two layers in the middle are called hidden layers. In case the neural network has more than one hidden layer it is called a deep neural network. A deep learning model can consist of several hidden layers and is trained with stochastic gradient descent and backpropagation (Goodfellow et al., 2016).

A standard neural network operation consists of multiplying the input features by a weight matrix and applying a non-linearity (activation function). Input variables $X_i = (X_1, X_2, \ldots, X_n)$ are fed into the neural network, weights $W_i = (W_1, W_2, \ldots, W_n)$ are added to each of the inputs and a linear combination of $\sum X_i W_i = w^T x$ is calculated. This linear

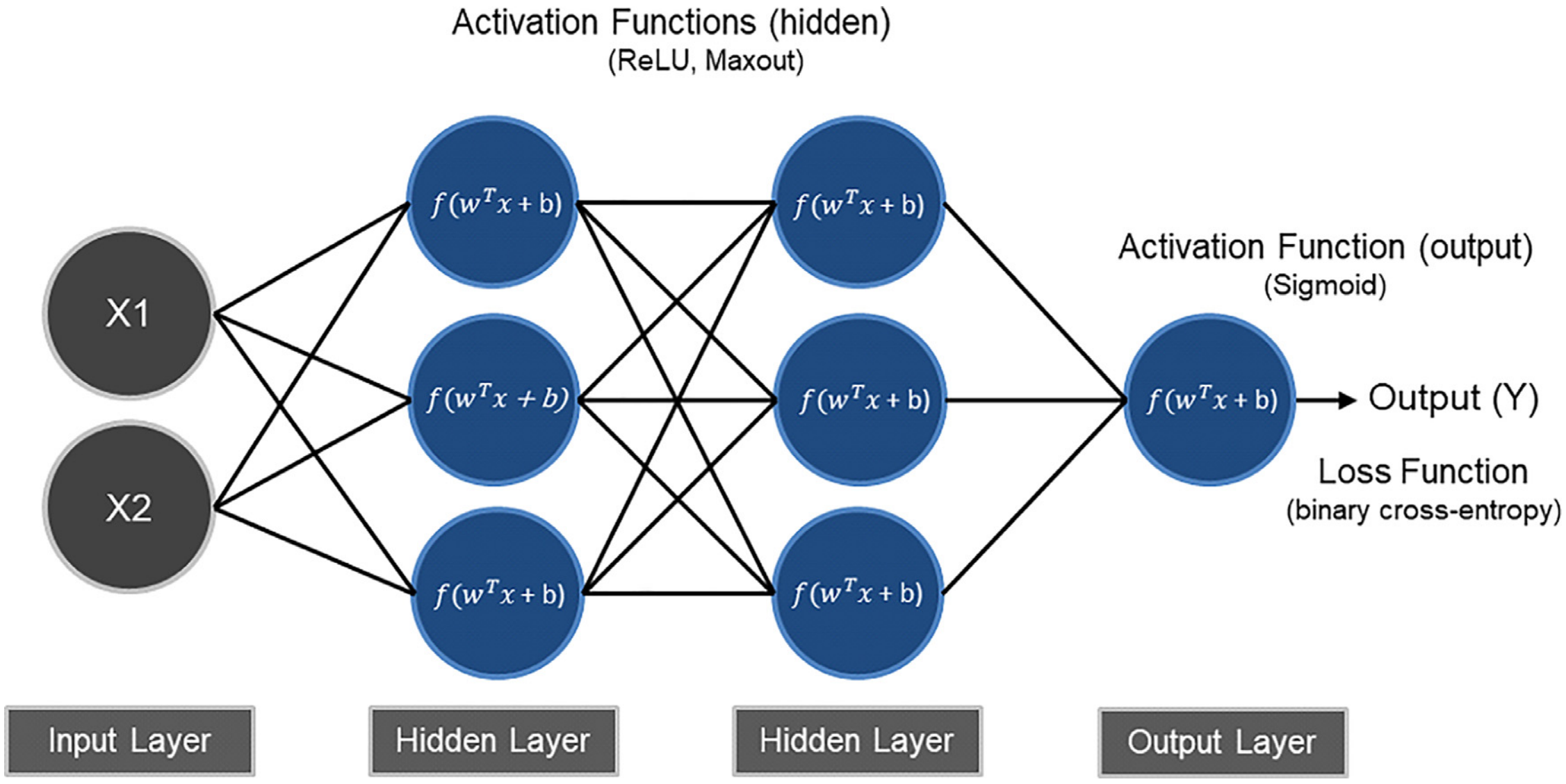

**Fig. 1.** The deep learning model used in this experiment is called a feedforward artificial neural network as the signal flow through the network evolves only in a forward direction. It is the most appropriate choice for problems based on structured datasets as used in this study. It contains one input as well as one output layer and various hidden layers. At each node, a linear combination of input variables and weights is fed into an activation function to calculate a new set of values for the next layer.

**Table 1**
Description of datasets.

| Business Area | Observations | | | | | Description |
|---|---|---|---|---|---|---|
| | Total | y = 0 | y = 1 | Balanced* | Features | |
| Credit Risk | 30,000 | 23,364 | 6,636 | 6636/6636 | 23 | Prediction whether a customer is going to default on their loan payment |
| Insurance Claims | 595,212 | 573,518 | 21,694 | 21694/21694 | 57 | Prediction whether a policy holder will initiate an auto insurance claim in the next year |
| Marketing/Sales | 45,211 | 39,922 | 5,289 | 5289/5289 | 16 | Prediction whether a targeted customer will open a deposit account after a direct marketing/sales effort |

*For the purpose of this study random under-sampling was used to bring the datasets in a balanced state

combination plus the bias term or interceptor serves as input for the activation function to calculate the output Y, which serves either as input for the next layer or represents the final output/prediction. A neural network is trained with stochastic gradient descent and backpropagation.

Applying a non-linearity in the form of an activation function is essential for neural networks to be able to learn complex (non-linear) representations of the input datasets. The activation function transforms the output at each node into a nonlinear function.

This study will build two different DL classifiers using the following activation functions for the hidden layers:

- The rectified linear unit (ReLU): $g(z) = \max(0, z) \in [0, \infty)$,
- The Maxout function: $g(z) = \max(w^k z + b^k) \in (-\infty, \infty),\ k \in \{1, \ldots, K\}$.

As the scope of the research is binary classification of structured data the output activation function used is the sigmoid function $sigm(z) = \frac{1}{1+e^{-z}} = \frac{e^z}{e^z+1} \in [0, 1]$ in line with the binary cross-entropy loss function.

## 2.2. Experimental design

### 2.2.1. Data and preprocessing

This experiment is based on three datasets. All three use cases require the same ML method, which is supervised learning and binary classification, and were used in earlier studies, which allows for easy comparison of classifier strength regarding earlier studies. To facilitate reproducibility and comparability the chosen data sets are all publicly available and can either be downloaded from the UCI machine learning repository or from the public machine learning competition site “Kaggle”, which regularly offers access to high-quality datasets for experimentation. See Table 1 for an overview of the case studies/datasets used in this study.

*Credit risk.* The first dataset represents payment information from Taiwanese credit card clients. It consists of 30,000 observations, of which 23,364 are good cases and 6,636 are bad cases (flagged as defaults). Each observation contains 23 features including a binary response column for the default information of the credit cardholder. The features within the dataset contain mainly historical payment information, but also demographic information such as gender, age, marital status, and education. [1]

*Insurance claims.* The second dataset represents information about automotive insurance policyholders. It consists of 595,212 observations, of which 573,518 are non-filed and 21,694 are filed claims. Each observation contains 57 features including a binary response column that indicates whether or not a particular policyholder has filed a claim. [2]

*Marketing and sales.* The third dataset stems from a retail bank and represents customer information for a direct marketing campaign. It consists of 45,211 observations, of which 39,922 were unsuccessful and 5,289 were successful (resulting in a sale). Each observation contains 16 features including a binary response column indicating whether or not the person ended up opening a deposit account with the bank following the direct marketing effort. [3]

The experiment required several adjustments. All three datasets are highly unbalanced. For this study, random under-sampling was used to bring the good as well as the bad cases into a state of equilibrium. This can also be seen in Table 1. Example: If highly unbalanced datasets with a ratio of 90:10 are trained it is very easy for the classifier to reach an accuracy of 90% by simply going for the positive observations in all cases. To counter this naturally occurring gravitation towards the majority class resampling is used to better gauge the predictive ability of the classifiers. One drawback of under-sampling might be a loss of information, but can be neglected as the major purpose of the dataset is to benchmark the introduced ML classifiers.

Before model construction can take place, several other common preprocessing steps have been performed. A required procedure in ML during preprocessing is to transform categorical values into a numerical representation. Especially the “Case Study 3 – Marketing and Sales” contains predominately categorical strings. Where necessary categorical features were transformed into factor variables with a method called one-hot encoding. H2O has a parameter setting called one_hot_explicit, which creates N+1 new columns for categorical features with N levels.

For this experimental study, all three datasets are separated into a training set and a test set with a proportion of 80:20. To tune the model parameters, the training set will be further divided into different training and validation sets using a method called cross-validation during the construction of the classifiers. Cross-validation is used to increase the generalization ability of the classifiers to unknown data and to avoid overfitting. This study uses 5-fold cross-validation.

Model tuning in ML is a highly empirical and interactive process and is essentially based on trial and error. The methods commonly used to help with automating the model tuning process are grid search and random search. Grid search automatically trains several models with different parameter settings over a predefined range of parameters. Overall, this does not change the basic necessity of trying out different combinations of parameters that allow the classifier to adjust adequately to the underlying dataset. This study used a random search, selective grid search, and manual adjustments to arrive at the final parameter settings.

The four performance evaluation measures (Flach, 2019) used in this study are AUC, Accuracy, F-score, and LogLoss.

### 2.2.2. Software

Data preparation and handling are managed in RStudio, which is the integrated development environment (IDE) for the statistical programming language R. R is one of the go-to languages for Data Science research as well as prototyping in practice. The machine learning models in this paper are developed with H2O, which is an open-source machine learning platform written in Java and supports a wide range of predictive models (LeDell & Gill, 2019). This makes experimentation and research easier. The high abstraction level allows the idea and the data to become the central part of the problem and helps to reduce the effort required to reach a solution. Also, H2O has the advantage of speed as it allows us to move from a desktop- or notebook-based environment to a large-scale environment. This increases performance and makes it easier to handle large data sets. R is connected to H2O by means of a REST API.

[1] The “Credit Risk” dataset can be accessed here: https://archive.ics.uci.edu/ml/datasets/default+of+credit+card+clients

[2] The “Insurance Claims” dataset can be accessed here: https://www.kaggle.com/c/porto-seguro-safe-driver-prediction/data

[3] The „Marketing/Sales dataset can be accessed here: https://archive.ics.uci.edu/ml/datasets/Bank+Marketing

## 3. Numerical results

In this section, three different case studies: Credit risk, insurance claims, and marketing and sales are presented to demonstrate that deep learning while being promoted as a superior ML solution has difficulties beating traditional machine learning methods in some cases. Concrete, logistic regression, random forest, gradient boosting machine, and two different deep learning classifiers were trained on each dataset. The first DL model was built with the ReLU activation function whereas the second DL model was built with the Maxout activation function. The ReLU activation function is widely used and has shown to be superior in terms of accuracy and computational speed. The Maxout activation function has been developed to improve classification accuracy in combination with dropout (Goodfellow, Warde-Farley, Mirza, Courville, & Bengio, 2013; Srivastava, Hinton, Krizhevsky, Sutskever, & Salakhutdinov, 2014) and is hence the second choice for this experiment. Several hyper-parameters were adjusted during the model training process to improve the performance measured by the evaluation metrics AUC, Accuracy, F-score, and LogLoss.

### 3.1. Case study 1: credit risk

Numerical results for the credit risk business case to accurately predict the default category of an applicant. The performance of deep learning is compared to traditional machine learning classifiers via the four evaluation matrices AUC, Accuracy, F-score, and LogLoss. The best performance is highlighted in bold.

Table 2 shows clearly that GBM has the best overall performance with the highest AUC, Accuracy, and F-score of 0.774, 0.712, and 0.691, respectively, including a LogLoss of 0.572. RF comes as a close second with an AUC of 0.773 and the same LogLoss as GBM of 0.572. Both ensemble models achieve a better performance in the case of the credit risk dataset than the two DL models with an AUC of 0.760 and 0.762, respectively. The DL + Maxout model has a slightly higher AUC compared to the DL + ReLU, whereas the LogLoss is reversed, which results in a similar performance for the two DL models.

A graphical presentation of the results of each model sorted by the evaluation measure can be found in Fig. 2. The best-performing model GBM is highlighted via a callout text field, which shows the performance of each evaluation metric.

**Table 2**
Numerical results for case study 1 - credit risk.

| Method | Out-of-Sample Performance | | | |
|---|---|---|---|---|
| | AUC | Accuracy | F-score | Logloss |
| Logistig Regression | 0.712 | 0.671 | 0.653 | 0.623 |
| Random Forest | 0.773 | 0.711 | 0.688 | 0.572 |
| **Gradient Boosting Machine** | **0.774** | 0.712 | 0.691 | 0.572 |
| Deep Learning + ReLU | 0.760 | 0.700 | 0.646 | 0.592 |
| Deep Learning + Maxout | 0.762 | 0.703 | 0.687 | 0.599 |

**Table 3**
Numerical results for case study 2 - insurance claims.

| Method | Out-of-Sample Performance | | | |
|---|---|---|---|---|
| | AUC | Accuracy | F-score | Logloss |
| Logistig Regression | 0.629 | 0.594 | 0.586 | 0.667 |
| Random Forest | 0.636 | 0.598 | 0.584 | 0.667 |
| **Gradient Boosting Machine** | **0.640** | 0.602 | 0.588 | 0.664 |
| Deep Learning + ReLU | 0.628 | 0.597 | 0.540 | 0.670 |
| Deep Learning + Maxout | 0.633 | 0.597 | 0.534 | 0.669 |

### 3.2. Case study 2: insurance claims

In Table 3 the numerical results for the insurance case study are presented. The goal is to accurately predict whether a policyholder is going to file an insurance claim within the next year. The performance of deep learning is compared to traditional machine learning classifiers via the four evaluation matrices AUC, Accuracy, F-score, and LogLoss. The best performance is highlighted in bold.

The results of Table 3 are similar to the first case study. GBM is the clear winner in terms of performance with the highest AUC, Accuracy, and F-score of 0.640, 0.602, and 0.588, respectively, including the lowest LogLoss of 0.664. RF takes second place with an AUC of 0.773 and a LogLoss of 0.664. Both ensemble models achieve a better performance in the insurance case study than the two DL models. The DL + Maxout model with an AUC of 0.633 has a slightly higher AUC compared to the DL + ReLU with an AUC of 0.628.

A graphical presentation of the results of each model sorted by the evaluation measure can be found in Fig. 3. The best-performing model (Gradient Boosting) is highlighted via a callout text field.

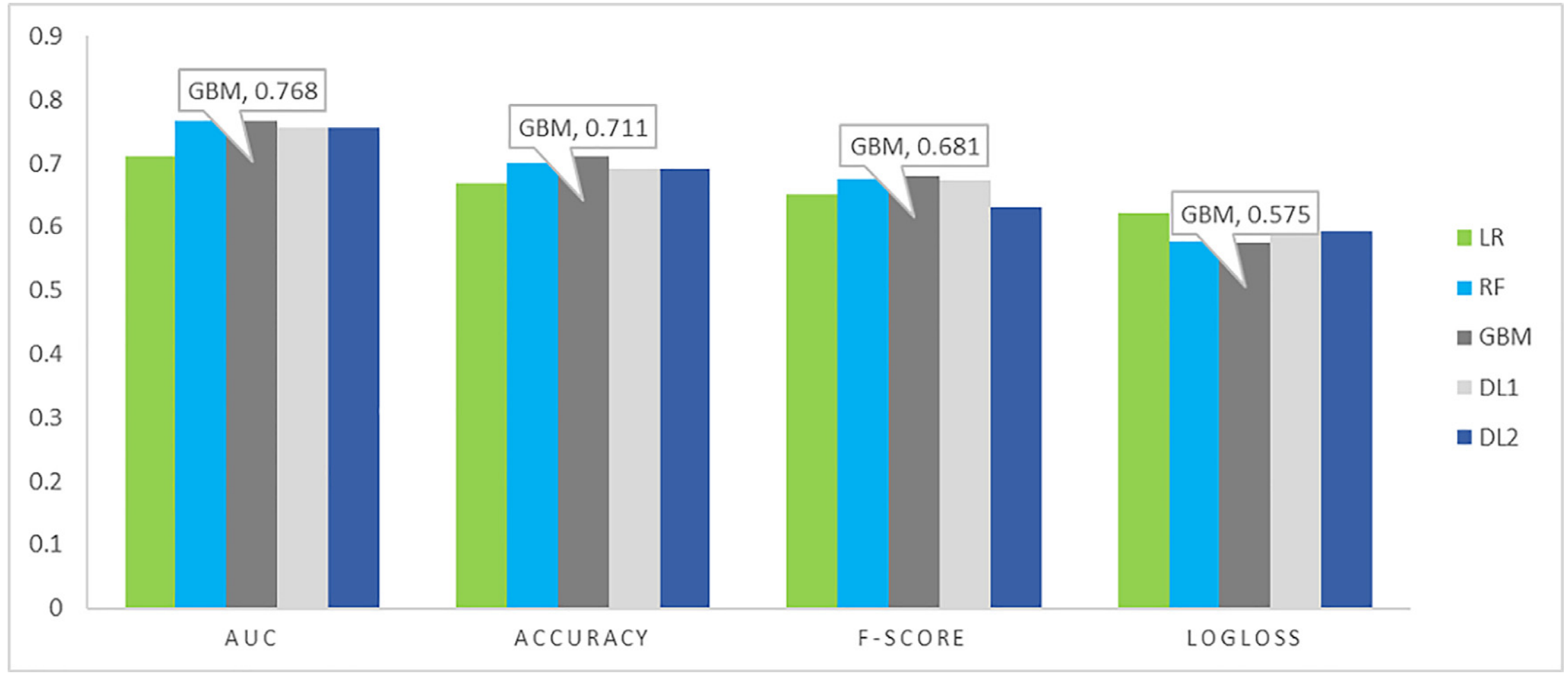

**Fig. 2.** Graphical representation of the performance of each classifier for all 4 performance evaluation metrics for case study 1 - credit risk. Gradient Boosting Machine (GBM) achieves the highest accuracy according to those results.

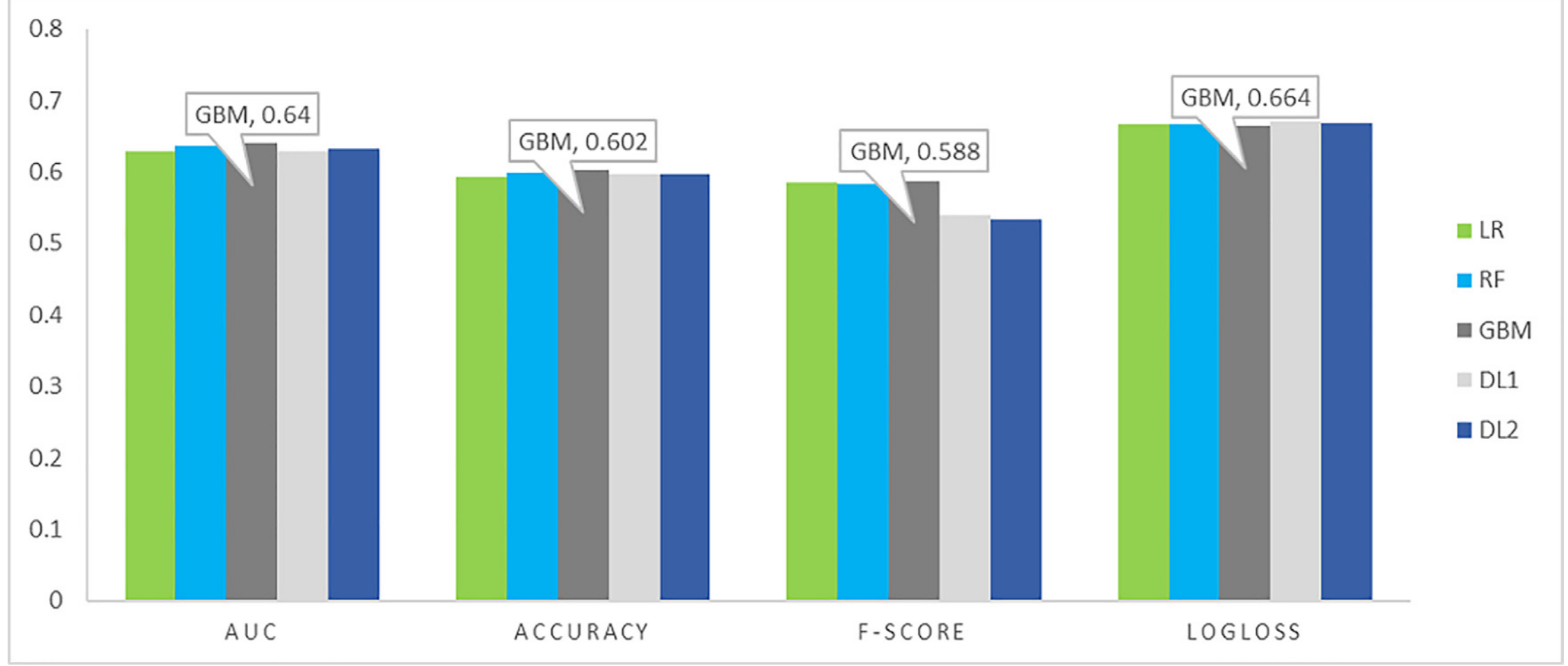

**Fig. 3.** Graphical representation of the performance of each classifier on all 4 performance measures for case study 2 - insurance claims. Also, in the second case study, Gradient Boosting Machine (GBM) achieves the highest prediction accuracy.

**Table 4**
Numerical results for case study 3 - marketing and sales.

| Method | Out-of-Sample Performance | | | |
|---|---|---|---|---|
| | AUC | Accuracy | F-score | Logloss |
| Logistig Regression | 0.918 | 0.839 | 0.845 | 0.377 |
| Random Forest | 0.940 | 0.879 | 0.888 | 0.320 |
| **Gradient Boosting Machine** | **0.940** | 0.878 | 0.886 | 0.299 |
| Deep Learning + ReLU | 0.930 | 0.861 | 0.877 | 0.328 |
| Deep Learning + Maxout | 0.930 | 0.857 | 0.865 | 0.336 |

### 3.3. Case study 3: marketing and sales

Table 4 shows the numerical results for the marketing and sales case study to accurately predict successful conversions based on a direct marketing effort. The performance of deep learning is compared to traditional machine learning classifiers via the four evaluation metrics AUC, Accuracy, F-score, and LogLoss. The best performance is highlighted in bold.

Based on Table 4 the results for the third case study are slightly different from case studies one and two. GBM shares the maximum AUC of 0.940 with RF. The RF classifier has also a slightly higher Accuracy of 0.879, and also a higher F-score of 0.888 while GBM has still the lowest LogLoss, which indicates the highest prediction reliability across the models. In line with previous results, both ensemble models achieve a better performance than the two DL models, which have both an AUC of 0.930. LR underperforms all classifiers by a significant amount.

A graphical presentation of the results of each model clustered by the evaluation measure can be found in Fig. 4. GBM and RF perform better than the two DL models across all performance measures while logistic regression turns out the be the weakest classifier.

## 4. DL in business analytics: a reality check

### 4.1. Discussion of results

To better understand the utility of Deep Learning for Business Analytics it was benchmarked against traditional ML models such as GLMs, Random Forest, and Gradient Boosting Machine. Based on the four evaluation measures AUC, Accuracy, F-score, and LogLoss.

The empirical results of all three case studies presented (Credit Risk, Insurance Claims, Marketing and Sales) suggest that DL does not have a performance advantage for classification problems based on structured data. Instead, the results are strongly in favor of tree-based ensembles such as random forest and gradient boosting. GBM turns out to be the model with the highest utility for the type of problems analyzed in this study.

Kraus et al. (2019) benchmarked several baseline models against their proposed embedded DNN model, which resulted in superior performance for DL. The authors recommend fostering the adoption of DL models within the field of Business Analytics and operations research. While the paper of Kraus et al. (2019) is an excellent overview of DL for Business Analytics and is very insightful, the analysis does not include GBM as a baseline model in the comparison, which is widely used and known to deliver strong and robust predictions on structured datasets.

Case study two in this study uses the same dataset as Kraus et al. (2019) and according to the empirical results is GBM at least on par with the proposed deep architecture by Kraus et al. (2019). Other studies by Hamori et al. (2018) and Addo et al. (2018) included tree-based ensembles as gradient boosting and came to the same conclusions as this study. As the findings of this study are in line with several papers comparing the performance of DL against other ML models there is strong evidence that tree-based methods (GBM as well as Random Forest) do indeed outperform DL models (different configurations have been tested) on most problems containing structured data. Also, DL has several weaknesses such as computational complexity, huge data requirements, transparency issues, and needs highly skilled labor, which makes it often difficult to develop and deploy those models at scale. Especially the computational complexity issue results in significantly longer training and validation times compared to all other ML models.

### 4.2. Contributions to literature

RQ1: Does DL outperform traditional ML models for supervised learning problems in the case of structured data with fixed-length feature vectors?

The empirical results suggest that deep learning does not have a performance advantage for classification problems based on structured datasets with fixed-length feature vectors. The results are strongly in favor of tree-based ensembles such as random forest and gradient boosting. These results strengthen the findings of earlier studies (Addo et al., 2018; Hamori et al., 2018; Schmitt, 2022b) which were predominately focusing on applications within credit risk management. This paper has extended the application domain with insurance, marketing, and sales use cases and it was shown that the outperformance of GBM for structured datasets is not an isolated phenomenon restricted to a single do-

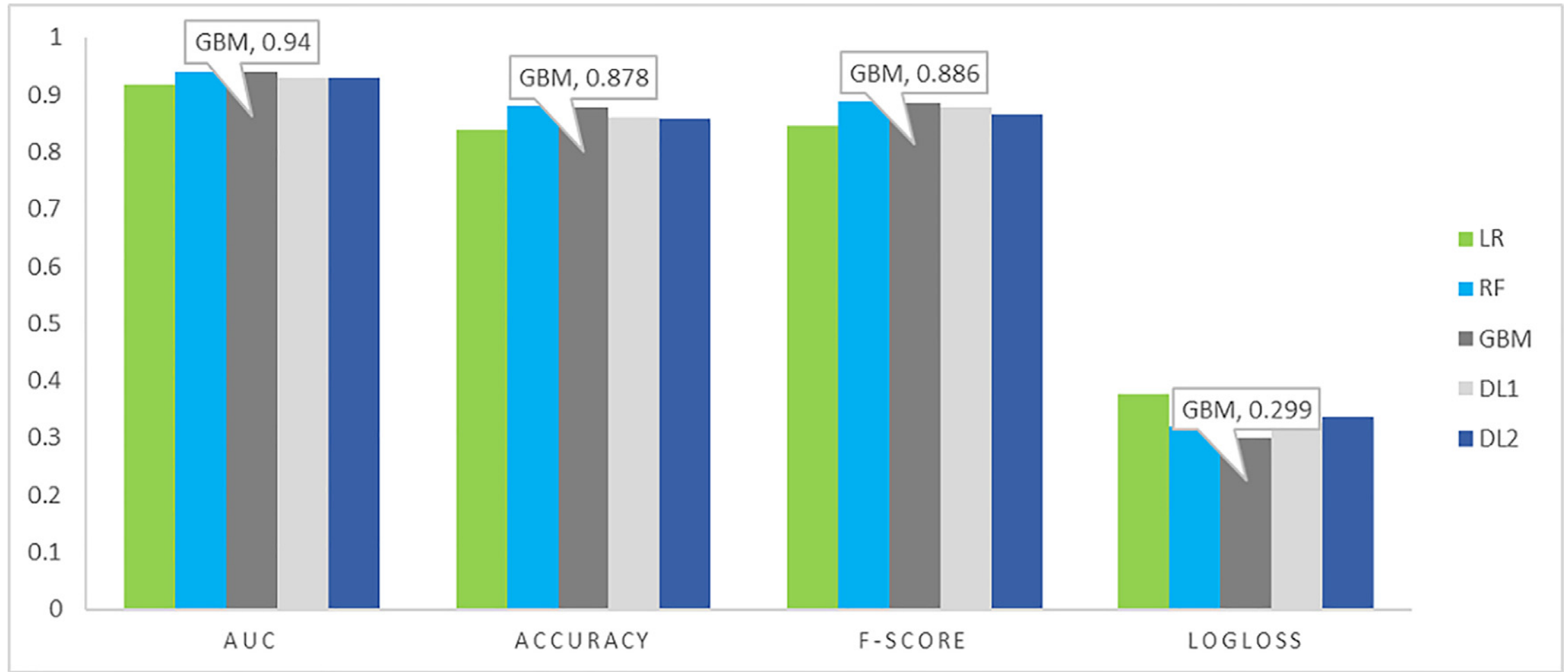

**Fig. 4.** Graphical representation of the performance of each classifier on all 4 performance measures for case study 3 – marketing and sales. Gradient Boosting Machine (GBM) is again the winner, but the results are less significant than before, and Random Forest (RF) achieves a very similar performance.

main – it is a fact that can be generalized across different fields that depend on business analytics and information management for powerful data-driven decision-making.

RQ2: Is Deep Learning – despite its popularity – always the right AI/ML model within business analytics and information management?

Deep learning should be regarded as a powerful addition to the existing body of ML models instead of a "one size fits all" solution. Earlier studies (Chui et al., 2018; Grover, Kar, & Dwivedi, 2022; Kar et al., 2021; Samek & Müller, 2019) have identified different barriers to AI adoption as computational complexity, missing big data infrastructure, lacking transparency, skill shortage, leadership commitment, and missing strategic guidance. All of those findings have the potential to impact the adoption speed of DL across different domains. In addition, the findings in this paper reveal that the prediction accuracy of DL is not always superior compared to other ML models. The results strongly suggest that gradient boosting can be seen as the go-to model for most business analytics problems. It is fast, not too complex, and delivers for use cases based on structured data the best performance currently available. The results are clear, however, business analytics experts should carefully consider the type, characteristics, and volume of the data at hand to make a final decision about the correct model choice. This is an important overall conclusion and an additional factor that impacts the adoption of DL for data-driven decision-making in business analytics and information management.

### *4.3. Implications for practice*

It has been proven that data-driven or evidence-based decisions are superior compared to pure intuitive business decisions and a comprehensive analytics strategy has become necessary for businesses across all industries to capture value at the bottom line. One of the challenges associated with becoming a digital enterprise is how exactly to leverage digital technologies and especially advanced analytics and AI. Current discussions about AI and digital strategy are strongly focused on the applications of DL, but this is not the best way to approach digital transformation. This focus resulted in the problematic assumption that DL adoption in business by itself can be regarded as a benchmark – thereby ignoring the question of utility that always needs to be asked before the deployment of any new method or technology.

The main explanation why DL has not found its way into the different business functions as expected is often explained by computational complexity, lacking big-data infrastructure, the non-transparent nature of DL (black-box), and a shortage of skills. But as was demonstrated in this paper, an additional explanation for the lack of adoption in certain business analytics functions is that DL does not have performance advantages over traditional analytics when it comes to structured data use cases.

For example, many departments that have been utilizing advanced analytics as risk management are perfectly capable of developing and deploying a DL model as the required skillset is identical. Also, the necessary infrastructure to leverage DL in these departments should be in place. The usually described problems are not the only reasons. The problem is that DL does not offer any advantage over certain tree-based ensembles for the data present in those departments. Also, the disadvantages of speed and transparency are still present, which makes it, in fact, unreasonable to use DL instead of traditional analytics. DL should be viewed as a valuable addition to the current body of ML that offers the possibility to create new use cases based on its strength instead of forcefully replacing models that are equally powerful and can easily coexist within advanced analytics.

This realization triggers the second argument, which is related to the nature of the underlying dataset. The kind of data present in problems faced within business analytics can largely be divided into three groups (Chen et al., 2012): (1) Structured data from relational database management systems (DBMS), (2) unstructured data, which stem mainly from web-based activities (Social Media Analytics, etc.), and (3) sensor- and mobile-based content, which is largely untouched when it comes to research activities. Many problems in business analytics are indeed based on structured datasets and given that most business functions utilize exactly those kinds of data it should not come as a surprise that DL remains a rather scarce ML algorithm to support their decision-making.

The era of big data has brought tremendous amounts of data within a single data set across several domains, which fulfills the requirement of empirical prediction based on deep learning. However, it is important to differentiate and use DL models mainly in line with their strength, which is the usage of vast unstructured datasets, which posed significant problems for traditional analytics. ML overall has been recognized as a general-purpose technology (GPT) for decision-making, which has just started to infuse our economy with the ability to replace mental tasks that were traditionally only reserved for humans (Agrawal et al., 2019). It has also the potential to create completely new business models (Siebel, 2019). Finding use cases that are in line with the strength of DL would help to foster the adoption of DL in business analytics. And the major strength is unprecedented accuracy on unstructured datasets. Traditional ML models reach a performance plateau quite early and further data are not helpful to increase accuracy. DL has here an advantage as it gains predictive power with every additional data point (Ng, 2019). This makes DL extremely scalable and future-proof, especially since hardware power and the amount of available data will increase continuously over the years. Also, DL eliminates the need for extensive feature engineering as this was usually present in the preprocessing stage of data mining and predictive analytics tasks (LeCun et al., 2015). The time required for preparing data sets often amounts to 80% to 90% of overall task completion and is one of the major reasons why further advances in DL would indeed be welcoming news for all analytics functions. Overall, management and practitioners responsible for digital strategy and transformation should avoid seeing DL as a simple replacement or enhancement of existing tools for predictive analytics tasks, but as an opportunity to develop new application areas and use cases for business analytics based on the strength of DL – which are predictions based on vast amounts of unstructured data.

### *4.4. Future research*

The following four key areas could be identified where further research is necessary to increase the utility and hence the adoption of DL in business analytics.

(1) Future research in business analytics could focus on identifying currently non-existing uses which are in line with the strength of DL. Due to its ability to handle huge amounts of unstructured data DL is in terms of future possibilities and new use cases more interesting than traditional analytics. DL possesses the ability to create completely new business models and ways of value generation.
(2) Enhancing the prediction accuracy of DL for structured data would be a game-changing development for neural networks. DL has several advantages over traditional methods but has in its current capacity difficulties reaching the performance and accuracy levels of tree-based ensembles such as Random Forest and GBM for predictions on structured data. A simple replacement makes hence no sense unless further research in this area realizes performance improvements for DL on structured classification tasks. Developments such as dropout (Srivastava et al., 2014) and the Maxout activation function (Goodfellow et al., 2013), which were specifically developed to tackle classification problems are going in this direction, but as shown above, are not enough to reach accuracy levels to justify the replacement of tree-based ensemble models as RF or GBM. Further research could focus on enhancing the ability of DL models to consistently surpass traditional ML models. This would be a significant development, which could result in the extinction of all other ML models.
(3) Another issue – especially in light of the skill shortage – is that hyperparameter tuning can be a quite complex undertaking requiring the right talent. A recent development is automated machine learning or AutoML, which has started to gain traction and is an interesting field of research that can help to further democratize the use of DL models (Schmitt, 2022a). Increasing the user-friendliness of AI by decreasing complexity, and aligning it with the end user's needs to increase job fit will help to foster adoption (Grover et al., 2022). AI needs to adapt to humans to enable a fully augmented workforce.
(4) This study was primarily concerned with binary classification, hence an extension towards multiclass classification and regression would make sense. Especially regression is relevant for finance and insurance due to the presence of financial times series data in those fields. Several studies have shown that deep learning architectures such as recurrent neural networks (RNN) and long short-term memory (LSTM) are strong candidates for time series data in finance and offer superior performance (Fischer & Krauss, 2018).
(5) Other areas for investigation would be reinforcement learning applications within business analytics (Singh et al., 2022), bio-inspired computation/ML models (Jain, Batra, Kar, Agrawal, & Tikkiwal, 2022; Kudithipudi et al., 2022), and also research that would further enhance the explainability of AI/ML, which would enable additional use cases in regulatory environments that require transparency (Bücker, Szepannek, Gosiewska, & Biecek, 2022; Sharma, Kumar, & Chuah, 2021).

## 5. Conclusion

The progress and breakthroughs achieved by DL are undeniable as can be witnessed by a vast array of new real-world applications all around us. Despite this fact, the adoption rate and hence diffusion across business analytics functions has been lacking behind. This study helped to explain the current lack of adoption of DL in business analytics functions. The literature analysis suggested that the lack of adoption across business functions is based on the five bottlenecks computational complexity, no existing big-data architecture, lack of transparency/black-box nature of DL, skill shortage, and leadership commitment. However, the empirical study based on three real-world case studies revealed that DL does not offer – as widely assumed – a performance advantage when it comes to predictions based on structured data sets. This has to be taken into account when using deep learning for data-driven decisions within the context of business analytics and answers the question of why analytics departments do not deploy those models consistently. Overall, ML as a general-purpose technology for data-driven prediction will further find its way into business analytics and shape the field of information management. Deep learning is a valuable addition to the ML ecosystem and enhanced our ability to generate insights from unstructured data. But it is not yet possible to replace the other models. Especially tree-based models such as random forest and gradient boosting are powerful classifiers when it comes to structured datasets. Practitioners should concentrate on creating new use cases that leverage the strengths of DL instead of forcing the replacement of traditional models.

## Declaration of Competing Interest

The authors declare that they have no known competing financial interests or personal relationships that could have appeared to influence the work reported in this paper.

## References

Addo, P. M., Guegan, D., & Hassani, B. (2018). Credit risk analysis using machine and deep learning models. *Risks, 6*(2), 1–20. 10.3390/risks6020038.
Agrawal, A., Gans, J., & Goldfarb, A. (2019). *The economics of artificial intelligence: An agenda*. London: National Bureau of Economic Research (Agrawal, A., Gans, J., & Goldfarb, A., eds.).
Baesens, B., Bapna, R., Marsden, J. R., Vanthienen, J., & Zhao, J. L. (2016). Transformational issues of big data and analytics in networked business. *Mis Quartely, 40*(4), 807–818.
Bertsimas, D., & Kallus, N. (2019). From predictive to prescriptive analytics. *Management Science*. 10.1287/mnsc.2018.3253.
Borges, A. F. S., Laurindo, F. J. B., Spínola, M. M., Gonçalves, R. F., & Mattos, C. A. (2021). The strategic use of artificial intelligence in the digital era: Systematic literature review and future research directions. *International Journal of Information Management, 57*(December 2019), Article 102225. 10.1016/j.ijinfomgt.2020.102225.
Breiman, L. (1996). Bagging predictors. *Machine Learning*. 10.1007/bf00058655.
Bücker, M., Szepannek, G., Gosiewska, A., & Biecek, P. (2022). Transparency, auditability, and explainability of machine learning models in credit scoring. *Journal of the Operational Research Society, 73*(1), 70–90. 10.1080/01605682.2021.1922098.
Bughin, J., Hazan, E., Ramaswamy, S., Chui, M., Allas, T., Dahlström, P., … Trench, M. (2017). *Artificial intelligence: The next digital frontier?* McKinsey&Company - McKinsey global institute. 10.1016/S1353-4858(17)30039-9.
Candel, A., & LeDell, E. (2019). *Deep learning with H2O*. (A. Bartz, Ed.), H2O. Ai (6th ed.). Retrieved from http://h2o.ai/resources/.
Chen, H., Chiang, R. H. L., & Storey, V. C. (2012). Business intelligence and analytics: From big data to big impact. *MIS Quarterly: Management Information Systems, 36*(4), 1165–1188. 10.5121/ijdps.2017.8101.
Chui, M., Manyka, J., Mehdi, M., Henke, N., Chung, R., Nel, P., & Malhotra, S. (2018). Notes from hundrets of insights from the AI frontier use cases. *McKinsey Global Institute*. McKinsey&Company.
Collins, C., Dennehy, D., Conboy, K., & Mikalef, P. (2021). Artificial intelligence in information systems research: A systematic literature review and research agenda. *International Journal of Information Management, 60*(July), Article 102383. 10.1016/j.ijinfomgt.2021.102383.
Davenport, T. H. (2018). From analytics to artificial intelligence. *Journal of Business Analytics, 1*(2), 73–80. 10.1080/2573234x.2018.1543535.

Delen, D., & Ram, S. (2018). Research challenges and opportunities in business analytics. *Journal of Business Analytics, 1*(1), 2–12. 10.1080/2573234x.2018.1507324.

Duan, Y., Edwards, J. S., & Dwivedi, Y. K. (2019). Artificial intelligence for decision making in the era of Big Data – Evolution, challenges and research agenda. *International Journal of Information Management, 48*(February), 63–71. 10.1016/j.ijinfomgt.2019.01.021.

Fischer, T., & Krauss, C. (2018). Deep learning with long short-term memory networks for financial market predictions. *European Journal of Operational Research, 270*(2), 654–669. 10.1016/j.ejor.2017.11.054.

Flach, P. (2019). Performance evaluation in machine learning: The good, the bad, the ugly, and the way forward. In *Proceedings of the AAAI conference on artificial intelligence: 33* (pp. 9808–9814). 10.1609/aaai.v33i01.33019808.

Fujii, M., Sakaji, H., Masuyama, S., & Sasaki, H. (2022). Extraction and classification of risk-related sentences from securities reports. *International Journal of Information Management Data Insights, 2*(2), Article 100096. 10.1016/j.jjimei.2022.100096.

Goodfellow, I., Bengio, Y., & Courville, A. (2016). Deep learning. *Deep Learning*. 10.1016/B978-0-12-391420-0.09987-X.

Goodfellow, I., Warde-Farley, D., Mirza, M., Courville, A., & Bengio, Y. (2013). Maxout networks. In *Proceedings of the 30th international conference on machine learning, ICML 2013* (pp. 2356–2364).

Grover, P., Kar, A. K., & Dwivedi, Y. K. (2022). Understanding artificial intelligence adoption in operations management: Insights from the review of academic literature and social media discussions. Annals of Operations Research *(Vol. 308)*. US: Springer. 10.1007/s10479-020-03683-9.

Hamori, S., Kawai, M., Kume, T., Murakami, Y., & Watanabe, C. (2018). Ensemble learning or deep learning? Application to default risk analysis. *Journal of Risk and Financial Management, 11*(1), 12. 10.3390/jrfm11010012.

Hastie, T., Tibshirani, R., & Friedman, J. (2017). *The elements of statistical learning second edition*. Springer Springer.doi.org/111.

Henke, N., Bughin, J., Chui, M., Manyika, J., Saleh, T., Wiseman, B., & Sethupathy, G. (2016). *The age of analytics: Competing in a data-driven world*. McKinsey Global Institute. Retrieved from https://www.mckinsey.com/business-functions/mckinsey-analytics/our-insights/the-age-of-analytics-competing-in-a-data-driven-world.

Hinton, G. E., Osindero, S., & Teh, Y. W. (2006). A fast learning algorithm for deep belief nets. *Neural Computation, 18*(7), 1527–1554. 10.1162/neco.2006.18.7.1527.

Jain, R., Batra, J., Kar, A. K., Agrawal, H., & Tikkiwal, V. A. (2022). A hybrid bio-inspired computing approach for buzz detection in social media. *Evolutionary Intelligence, 15*(1), 349–367. 10.1007/s12065-020-00512-7.

Jordan, M. I., & Mitchell, T. M. (2015). Machine learning: Trends, perspectives, and prospects. *Science, 349*(6245), 255–260. 10.1126/science.aaa8415.

Kar, S., Kar, A. K., & Gupta, M. P. (2021). Modeling drivers and barriers of artificial intelligence adoption: Insights from a strategic management perspective. *Intelligent systems in accounting. Finance and Management, 28*(4), 217–238. 10.1002/isaf.1503.

Kraus, M., Feuerriegel, S., & Oztekin, A. (2019). Deep learning in business analytics and operations research: Models, applications and managerial implications. *European Journal of Operational Research*, 1–14. 10.1016/j.ejor.2019.09.018.

Kudithipudi, D., Aguilar-Simon, M., Babb, J., Bazhenov, M., Blackiston, D., Bongard, J., ... Siegelmann, H. (2022). Biological underpinnings for lifelong learning machines. *Nature Machine Intelligence, 4*(3), 196–210. 10.1038/s42256-022-00452-0.

Kumar, S., Kar, A. K., & Ilavarasan, P. V. (2021). Applications of text mining in services management: A systematic literature review. *International Journal of Information Management Data Insights, 1*(1), Article 100008. 10.1016/j.jjimei.2021.100008.

Kushwaha, A. K., Kar, A. K., & Dwivedi, Y. K. (2021). Applications of big data in emerging management disciplines: A literature review using text mining. *International Journal of Information Management Data Insights, 1*(2), Article 100017. 10.1016/j.jjimei.2021.100017.

LeCun, Y., Bengio, Y., & Hinton, G. (2015). Deep learning. *Nature, 521*(7553), 436–444. 10.1038/nature14539.

LeDell, E., & Gill, N. (2019). *H2O: R Interface for "H2O"*. R Package Retrieved November 22, 2019, from https://cran.r-project.org/web/packages/h2o/index.html .

Lee, K. F. (2018). *AI superpowers: China, Silicon Valley, and the new world order*. Houghton Mifflin.

Malohlava, M., & Candel, A. (2019). *Gradient boosting machine with H2O. H2O. ai* (7th ed.). H2O. Retrieved from http://h2o.ai/resources/%0Ahttp://h2o-release.s3.amazonaws.com/h2o/master/3805/docs-website/h2o-docs/booklets/GBMBooklet.pdf.

Murphy, K. P. (2012). *Machine learning: A probabilistic perspective*. Cambridge, Massachusetts; London, England: The MIT Press. 10.1038/217994a0.

Ng, A. (2019). *Machine learning yearning: Technical strategy for ai engineers in the era of deep learning*. Retrieved online at https://www.mlyearning.org.

Rawat, S., Rawat, A., Kumar, D., & Sabitha, A. S. (2021). Application of machine learning and data visualization techniques for decision support in the insurance sector. *International Journal of Information Management Data Insights*, (2), 1. 10.1016/j.jjimei.2021.100012.

Samek, W., & Müller, K.R. (2019). *Towards explainable artificial intelligence*. 10.1007/978-3-030-28954-6_1.

Schmitt, M. (2020). *Artificial intelligence in business analytics: Capturing value with machine learning applications in financial services*. University of Strathclyde. 10.48730/5s00-jd45.

Schmitt, M. (2022a). Automated machine learning: AI-driven decision making in business analytics. *ArXiv Preprint*. 10.48550/arXiv.2205.10538.

Schmitt, M. (2022b). Deep learning vs. Gradient boosting: Benchmarking state-of-the-art machine learning algorithms for credit scoring. *ArXiv Preprint*. 10.48550/arXiv.2205.10535.

Sharda, R., Delen, D., & Turban, E. (2017). *Business intelligence, analytics, and data science: A managerial perspective*. Pearson Education Limited Retrieved from www.pearsonglobaleditions.com.

Sharma, R., Kumar, A., & Chuah, C. (2021). Turning the blackbox into a glassbox: An explainable machine learning approach for understanding hospitality customer. *International Journal of Information Management Data Insights, 1*(2), Article 100050. 10.1016/j.jjimei.2021.100050.

Siebel, T. M. (2019). *Digital transformation: Survive and thrive in an era of mass extinction* (1st ed.). New York: Rosetta Books.

Singh, V., Chen, S., Singhania, M., Nanavati, B., & Gupta, A. (2022). How are reinforcement learning and deep learning algorithms used for big data based decision making in financial industries – A review and research agenda. *International Journal of Information Management Data Insights, 2*(2), Article 100094. 10.1016/j.jjimei.2022.100094.

Sounderajah, V., McCradden, M. D., Liu, X., Rose, S., Ashrafian, H., Collins, G. S., ... Darzi, A. (2022). Ethics methods are required as part of reporting guidelines for artificial intelligence in healthcare. *Nature Machine Intelligence, 4*(4), 316–317. 10.1038/s42256-022-00479-3.

Srivastava, N., Hinton, G., Krizhevsky, A., Sutskever, I., & Salakhutdinov, R. (2014). Dropout: A simple way to prevent neural networks from overfitting. *Journal of Machine Learning Research*.

Stadelmann, T., Amirian, M., Arabaci, I., Arnold, M., Duivesteijn, G. F., Elezi, I., ... Tuggener, L. (2018). Deep learning in the wild. ArXiv, 11081 LNAI, 17–38. 10.1007/978-3-319-99978-4_2.

Taddeo, M., McCutcheon, T., & Floridi, L. (2019). Trusting artificial intelligence in cybersecurity is a double-edged sword. *Nature Machine Intelligence, 1*(12), 557–560. 10.1038/s42256-019-0109-1.

Thorat, O., Parekh, N., & Mangrulkar, R. (2021). TaxoDaCML: Taxonomy based divide and conquer using machine learning approach for DDoS attack classification. *International Journal of Information Management Data Insights, 1*(2), Article 100048. 10.1016/j.jjimei.2021.100048.

Verma, S., Sharma, R., Deb, S., & Maitra, D. (2021). Artificial intelligence in marketing: Systematic review and future research direction. *International Journal of Information Management Data Insights, 1*(1), Article 100002. 10.1016/j.jjimei.2020.100002.

Votto, A. M., Valecha, R., Najafirad, P., & Rao, H. R. (2021). Artificial intelligence in tactical human resource management: A systematic literature review. *International Journal of Information Management Data Insights, 1*(2), Article 100047. 10.1016/j.jjimei.2021.100047.

Warner, K. S. R., & Wäger, M. (2019). Building dynamic capabilities for digital transformation: An ongoing process of strategic renewal. *Long Range Planning, 52*(3), 326–349. 10.1016/j.lrp.2018.12.001.

Young, Z., & Steele, R. (2022). Empirical evaluation of performance degradation of machine learning-based predictive models – A case study in healthcare information systems. *International Journal of Information Management Data Insights, 2*(1), Article 100070. 10.1016/j.jjimei.2022.100070.